%% file: main.tex
\documentclass[]{fairmeta}

\usepackage{microtype}
\usepackage{lmodern}

\usepackage[utf8]{inputenc}
\usepackage[T1]{fontenc}

\usepackage{amsfonts}
\usepackage{amsmath}
\usepackage{amssymb}
\usepackage{amsthm}

\usepackage{graphicx}
\usepackage{booktabs}
\usepackage{float}
\usepackage{wrapfig}
\usepackage{multirow}
\usepackage{tabularx}
\usepackage{makecell}

\usepackage{algorithm}
\usepackage{algorithmic}

\usepackage[most]{tcolorbox}
\usepackage{enumitem}
\usepackage{xcolor}
\usepackage{url}
\usepackage{hyperref}
\usepackage[capitalize,noabbrev]{cleveref}
\usepackage{nicefrac}
\usepackage{pifont}

\theoremstyle{plain}

\theoremstyle{definition}

\theoremstyle{remark}

\graphicspath{{figures/}}

\newcolumntype{C}{>{\centering\arraybackslash}X}

\newcommand{\system}{\textsc{AutoResearchClaw}}
\newcommand{\bench}{\textsc{ARC-Bench}}
\newcommand{\cmark}{\textcolor{green!60!black}{\ding{51}}}
\newcommand{\xmark}{\textcolor{red!70}{\ding{55}}}
\newcommand{\tmark}{\textcolor{orange!80}{$\sim$}}

\newcommand{\eg}{\textit{e.g.}}

\title{%
  \raisebox{-0.25\height}{\includegraphics[height=0.9\baselineskip]{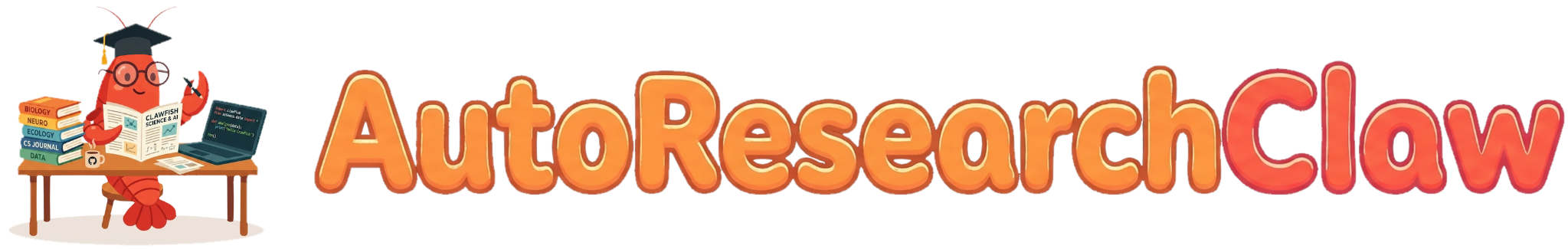}}%
  : Self-Reinforcing Autonomous Research with Human-AI Collaboration%
}

\author[1]{Jiaqi Liu$^*$}
\author[1]{Shi Qiu$^*$}
\author[1]{Mairui Li}
\author[1]{Bingzhou Li}
\author[1]{Haonian Ji}
\author[1]{Siwei Han}
\author[1]{Xinyu Ye}
\author[1]{Peng Xia}
\author[6]{Zihan Dong}
\author[1]{Meng Chen}
\author[1]{Congyu Zhang}
\author[2]{Letian Zhang}
\author[2]{Guiming Chen}
\author[2]{Haoqin Tu}
\author[3]{Xinyu Yang}
\author[1]{Lu Feng}
\author[7]{Xujiang Zhao}
\author[7]{Haifeng Chen}
\author[1]{Jiawei Zhou}
\author[8]{Xiao Wang}
\author[1]{Weitong Zhang}
\author[1]{Hongtu Zhu}
\author[1]{Yun Li}
\author[10]{Jieru Mei}
\author[10]{Hongliang Fei}
\author[4]{Jiaheng Zhang}
\author[11]{Linjie Li}
\author[6]{Linjun Zhang}
\author[2]{Yuyin Zhou}
\author[11]{Sheng Wang}
\author[12]{Caiming Xiong}
\author[9]{James Zou}
\author[5]{Zeyu Zheng}
\author[2]{Cihang Xie}
\author[1]{Mingyu Ding}
\author[1]{Huaxiu Yao}

\affiliation[1]{UNC-Chapel Hill}
\affiliation[2]{UC Santa Cruz}
\affiliation[3]{Carnegie Mellon University}
\affiliation[4]{NUS}
\affiliation[5]{UC Berkeley}
\affiliation[6]{Rutgers University}
\affiliation[7]{NEC Labs America}
\affiliation[8]{Meta}
\affiliation[9]{Stanford University}
\affiliation[10]{Google}
\affiliation[11]{University of Washington} 
\affiliation[12]{Recrusive.com}
\affiliation{\\$^*$Equal contribution. \quad Contact: \texttt{\{jqliu,shiqiu,huaxiu\}@cs.unc.edu}}

\abstract{
Automating scientific discovery requires more than generating papers from ideas.
Real research is iterative: hypotheses are challenged from multiple perspectives, experiments fail and inform the next attempt, and lessons accumulate across cycles.
Existing autonomous research systems often model this process as a linear pipeline: they rely on single-agent reasoning, stop when execution fails, and do not carry experience across runs.
We present \system{}, a multi-agent autonomous research pipeline built on five mechanisms: structured multi-agent debate for hypothesis generation and result analysis, a self-healing executor with a \textsc{Pivot}/\textsc{Refine} decision loop that transforms failures into information, verifiable result reporting that prevents fabricated numbers and hallucinated citations, human-in-the-loop collaboration with seven intervention modes spanning full autonomy to step-by-step oversight, and cross-run evolution that converts past mistakes into future safeguards.
On \bench, a 25-topic experiment-stage benchmark, \system{} outperforms AI Scientist v2 by 54.7\%.
A human-in-the-loop ablation across seven intervention modes reveals that precise, targeted collaboration at high-leverage decision points consistently outperforms both full autonomy and exhaustive step-by-step oversight.
We position \system{} as a research amplifier that augments rather than replaces human scientific judgment.
}

\metadata[Github]{\url{https://github.com/aiming-lab/AutoResearchClaw}}

\begin{document}

\maketitle

\input{sections/intro}
\input{sections/related_work}
\input{sections/system}
\input{sections/experiment}
\input{sections/analysis}
\input{sections/conclusion}

\bibliographystyle{plainnat}
\bibliography{references}

\clearpage
\newpage
\beginappendix

\input{sections/appendix}


\end{document}

%% file: sections/intro.tex
\vspace{-0.5em}
\section{Introduction}
\label{sec:intro}
\vspace{-0.5em}

Automating scientific discovery is a major goal of artificial intelligence.
Recent LLM-based systems have shown that agents can generate hypotheses, run experiments, and draft papers~\citep{lu2024aiscientist, yamada2025aiscientistv2, schmidgall2025agentlab, tang2025airesearcher}.
Real research, however, does not proceed in a straight line from idea to paper.
A researcher proposes a hypothesis, designs an experiment, observes what fails, revises the plan based on that failure, and tries again iteratively.
This loop depends on three capabilities: challenging one's own hypotheses from multiple angles, recovering from failed experiments without losing partial progress, and carrying lessons from past attempts into future ones.

Existing systems handle each of these capabilities poorly.
On hypothesis quality, single-agent systems such as AI Scientist~\citep{lu2024aiscientist, yamada2025aiscientistv2} use the same model to generate and evaluate hypotheses, which makes it harder to surface weak assumptions or overly easy directions.
On execution robustness, systems such as AIDE ML~\citep{aide2025} stop after an execution failure and discard partial results that could still be informative.
On experience accumulation, multi-agent systems such as Agent Laboratory~\citep{schmidgall2025agentlab} allow collaboration within a single run but do not carry lessons across runs, so each attempt starts from scratch.
The result is that research is treated as a one-off process rather than an iterative cycle.

Our key observation is that these three challenges are not independent.
Better hypotheses reduce the need for major revisions during execution.
More robust execution preserves intermediate results that can inform analysis.
Lessons from past runs can improve both hypothesis generation and experiment design in later attempts.
Improving one challenge therefore helps the others, which means they need to be addressed together in a unified framework.

We present \system{}, a multi-agent research pipeline built around five mechanisms that address these challenges jointly.
\textit{Structured multi-agent debate} assigns agents roles such as innovator, pragmatist, and contrarian, and has them critique each other during hypothesis generation and result analysis; a synthesizer then integrates their outputs into a single structured artifact.
\textit{A self-healing executor} uses a \textsc{Pivot}/\textsc{Refine} decision loop to treat failures as information rather than stopping points: after a failure, the system diagnoses the cause, then either adjusts the current experiment and retries (\textsc{Refine}) or moves to a new direction based on what the failure revealed (\textsc{Pivot}).
\textit{Verifiable result reporting} ties all reported numbers to a registry of executed outputs and checks every citation through a four-layer verification pipeline before anything appears in a draft.
\textit{Human-in-the-loop collaboration} provides seven intervention modes spanning full autonomy to step-by-step approval, with a confidence-driven SmartPause mechanism that routes decisions to the researcher only when system uncertainty is high.
\textit{Cross-run evolution} stores structured lessons from previous runs and injects them as guidance in future attempts through a time-decayed weighting scheme.
These mechanisms interact: past lessons inform debate, debate improves experiment choices, self-healing keeps the pipeline moving, and verification ensures outputs are grounded in actual results.

In summary, our main contribution is \system{}, an open-source multi-agent system for autonomous research that addresses hypothesis quality, execution robustness, and experience accumulation together.
We introduce \bench, a 25-topic benchmark focused on the experiment stage, evaluated with a rubric-assisted LLM judge.
On this benchmark, \system{} outperforms AI Scientist v2 by 54.7\%.
A human-in-the-loop ablation across seven intervention modes shows that targeted human input at high-leverage decision points consistently outperforms both full autonomy and dense step-by-step oversight.
Further analysis shows that the modular design of \system{} can connect to domain-specific scientific experiments, including high-energy theory.
We discuss safeguards for responsible use, including citation verification, claim grounding, and transparency requirements, in Appendix~\ref{sec:ethics}.

\begin{figure}[t]
\centering
\includegraphics[width=\linewidth]{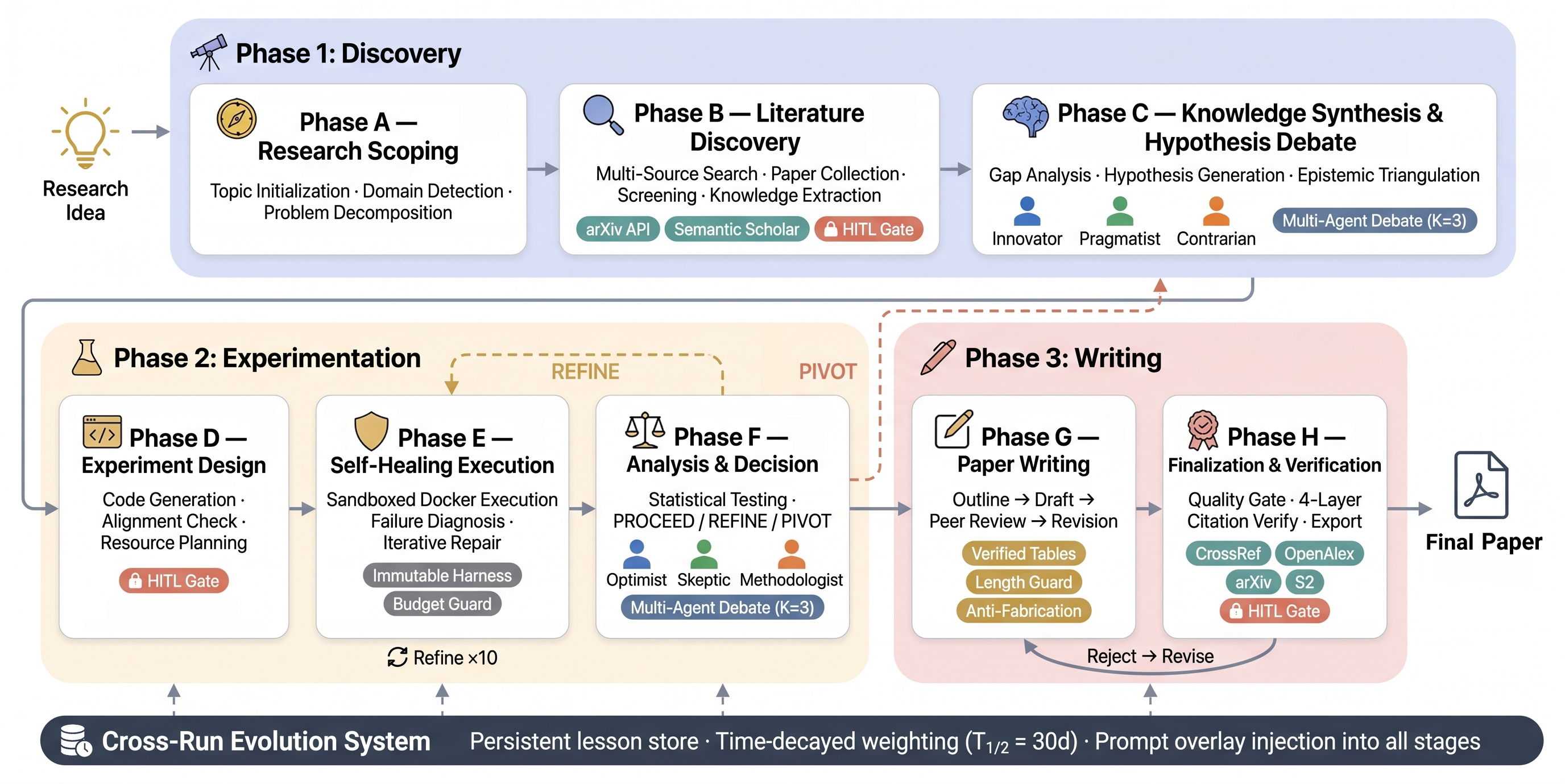}
\caption{Overview of the \system{} pipeline. Given a research idea, the 
system progresses through three stages: \textbf{Discovery} (scoping, 
literature search, multi-agent debate for hypothesis generation), 
\textbf{Experimentation} (self-healing code execution, result analysis with 
a second debate panel, and \textsc{Pivot}/\textsc{Refine} decisions), and 
\textbf{Writing} (drafting, review, revision, four-layer citation 
verification). Optional human-in-the-loop gates (orange) allow oversight 
at key checkpoints. The cross-run evolution system (bottom) injects 
time-decayed lessons from prior runs into all phases.}
\label{fig:pipeline}
\vspace{-1.5em}
\end{figure}
\vspace{-1em}

%% file: sections/related_work.tex
\section{Related Work}
\label{sec:related}
\vspace{-0.5em}

\noindent \textbf{Autonomous research systems.}
LLMs have been applied to autonomous experiment execution~\citep{boiko2023emergent} and algorithmic discovery~\citep{romera2024mathematical, novikov2025alphaevolve}. End-to-end research systems vary in scope and capability. The AI Scientist~\citep{lu2024aiscientist} and its successor~\citep{yamada2025aiscientistv2} generate complete papers from ideas but rely on single-agent reasoning, abort on execution failures, and start each run from scratch. AI Co-Scientist~\citep{gottweis2025aicoscientist, gottweis2025towards} introduces multi-agent debate for hypothesis validation but does not execute experiments. Agent Laboratory~\citep{schmidgall2025agentlab} and AI-Researcher~\citep{tang2025airesearcher} automate portions of the pipeline but neither verifies results against ground-truth measurements nor accumulates knowledge across runs. MLR-Copilot~\citep{li2024mlrcopilot} targets machine learning research with explicit human feedback at the execution stage. AgentRxiv~\citep{schmidgall2025agentrxiv} explores inter-agent collaboration through shared preprint servers. On the evaluation side, ScienceAgentBench~\citep{tian2024scienceagentbench}, MLE-bench~\citep{chan2024mlebench}, and DISCOVERYWORLD~\citep{jansen2024discoveryworld} reveal that even the best systems solve fewer than 40\% of tasks. As summarized in Table~\ref{tab:comparison}, no prior system combines end-to-end execution with multi-agent debate, self-healing, anti-fabrication verification, and cross-run evolution.

\noindent \textbf{Multi-agent debate and cross-run learning.}
Multi-agent debate improves factual accuracy and divergent thinking~\citep{du2023improving, liang2023encouraging, tran2025multiagent}. Role-assigned frameworks such as ChatDev~\citep{qian2024chatdev}, MetaGPT~\citep{hong2024metagpt}, and AutoGen~\citep{wu2024autogen} demonstrate effective collaboration in software engineering. For learning from experience, Reflexion~\citep{shinn2023reflexion} and Self-Refine~\citep{madaan2023selfrefine} operate within a single episode; SkillRL~\citep{xia2026skillrl} and EvolveR~\citep{wang2025evolver} extend this to persistent skill libraries across tasks. OmniScientist~\citep{shao2025omniscientist} argues that science is inherently collaborative and proposes protocols for multi-agent research ecosystems. \system{} applies debate with domain-specific epistemic roles at two pipeline stages and accumulates lessons \emph{across} runs through a persistent time-decayed store, combining both mechanisms in a single system.

\noindent \textbf{Human-AI collaboration in research automation.}
The degree of human involvement in autonomous research remains an open design question. At one extreme, the AI Scientist pursues full automation with minimal human oversight. At the other, SciSciGPT~\citep{shao2025sciscigpt} positions AI as an assistant under continuous human direction. Between these extremes, Agent Laboratory~\citep{schmidgall2025agentlab} allows user-defined feedback frequency and reports that human participation at each stage improves quality. AIssistant~\citep{gaddipati2025aissistant} demonstrates 65.7\% time savings through strategic human oversight in review writing. \citet{natarajan2024hitl} provide a theoretical analysis arguing that the optimal level of human intervention depends on how well-defined the task is. Our HITL ablation contributes empirical evidence to this debate: across seven intervention regimes, we find that targeted intervention at high-leverage decision points consistently outperforms both full autonomy and exhaustive step-by-step oversight.

\begin{table}[t]
\centering
\caption{Feature comparison of autonomous research systems. \cmark{} = supported, \xmark{} = not supported, \tmark{} = partial. Only AI Scientist v2 and \system{} provide end-to-end autonomous execution from idea to paper; Agent Laboratory and ResearchAgent stop short.}
\label{tab:comparison}
\setlength{\tabcolsep}{4pt}
\small
\begin{tabularx}{\textwidth}{@{}lCCCCCC@{}}
\toprule
\textbf{Feature}
  & \textbf{\thead{AI Sci.\\ v2}}
  & \textbf{\thead{AI\\ Co-Sci.}}
  & \textbf{\thead{Agent\\ Lab}}
  & \textbf{\thead{MLR-\\ Copilot}}
  & \textbf{\thead{Research-\\ Agent}}
  & \textbf{\thead{Auto\\ Research\\ Claw}} \\
\midrule
End-to-end pipeline        & \cmark & \tmark & \xmark & \tmark & \xmark & \cmark \\
Real experiment exec.\     & \cmark & \xmark & \tmark & \cmark & \xmark & \cmark \\
Multi-agent debate         & \xmark & \cmark & \xmark & \xmark & \xmark & \cmark \\
Self-healing repair        & \xmark & \xmark & \xmark & \xmark & \xmark & \cmark \\
\textsc{Pivot}/\textsc{Refine} loop & \xmark & \tmark & \xmark & \xmark & \xmark & \cmark \\
Cross-run evolution        & \xmark & \cmark & \xmark & \xmark & \xmark & \cmark \\
Citation verification      & \xmark & \xmark & \xmark & \xmark & \xmark & \cmark \\
Result verification        & \xmark & \xmark & \xmark & \xmark & \xmark & \cmark \\
HITL gates                 & \xmark & \xmark & \cmark & \cmark & \xmark & \cmark \\
Sandbox security           & \xmark & \xmark & \xmark & \xmark & \xmark & \cmark \\
Open-source                & \cmark & \xmark & \cmark & \cmark & \cmark & \cmark \\
\bottomrule
\end{tabularx}
\end{table}

%% file: sections/system.tex
\vspace{-0.5em}
\section{\system{}}
\label{sec:method}
\vspace{-0.5em}

\vspace{-0.3em}
\subsection{Overview}
\label{sec:overview}
\vspace{-0.3em}

\system{} is organized as a 23-stage pipeline across three phases (Figure~\ref{fig:pipeline}):
\textbf{Discovery} (scoping, literature search, multi-agent hypothesis generation),
\textbf{Experimentation} (self-healing code execution, result analysis, autonomous \textsc{Pivot}/\textsc{Refine} decisions),
and \textbf{Writing} (drafting, multi-agent review, revision, citation verification).
Five mechanisms span all three phases.
Multi-agent debate stress-tests hypotheses and conclusions from complementary perspectives.
Self-healing execution treats experiment failures as diagnostic information rather than termination signals.
Verifiable result reporting enforces that only grounded numbers and verified citations reach the final output.
Human-in-the-loop (HITL) collaboration allows researchers to intervene at high-leverage decision points without managing the full pipeline.
Cross-run evolution converts past failures into reusable safeguards through a persistent, time-decayed lesson store.
Each stage declares a formal input/output contract and supports checkpoint-based resumption; full stage definitions and hardware adaptation details are in Appendix~\ref{app:stages}.

\vspace{-0.3em}
\subsection{Multi-Agent Debate}
\label{sec:debate}
\vspace{-0.3em}

A single LLM agent naturally tends to confirm the hypotheses it generates, because the same model that proposes an idea has no structural incentive to disconfirm it.
\system{} addresses this by instantiating structured debate at two pipeline stages.
Each debate panel uses $K{=}3$ agents with complementary epistemic roles and a synthesizer that integrates their outputs into a single structured artifact.

\noindent \textbf{Hypothesis-stage debate.}
During hypothesis formulation, an \emph{Innovator} proposes high-risk hypotheses that challenge conventional assumptions, a \emph{Pragmatist} evaluates feasibility given hardware and time budgets, and a \emph{Contrarian} actively seeks weaknesses and confounds.
The synthesizer distills these perspectives into 2--4 falsifiable hypotheses, each annotated with testability criteria and required baselines.

\noindent \textbf{Result-stage debate.}
After experiments complete, a second panel evaluates the results.
An \emph{Optimist} surfaces strong findings, a \emph{Skeptic} challenges statistical significance and flags potential confounds, and a \emph{Methodologist} evaluates reproducibility and checks for data leakage.
The synthesizer produces a structured assessment that distinguishes supported claims from unsupported ones before any writing stage begins.

\vspace{-0.3em}
\subsection{Self-Healing Execution}
\label{sec:selfheal}
\vspace{-0.3em}

Experiment failure is common in real research.
Existing autonomous systems treat failure as a termination condition and discard all intermediate progress.
\system{} instead treats failure as diagnostic information: the system identifies what went wrong, decides whether to fix the current approach or change direction, and preserves all recoverable artifacts.

\noindent \textbf{Cascading code generation.}
Research experiments range from single-file scripts to multi-file systems with custom architectures.
A scoring function rates each experiment plan along six dimensions: architectural depth, file count, domain difficulty, dependency chains, historical failure rate, and control-flow complexity, and produces a complexity scalar $c \in [0,1]$.
Experiments above a fixed threshold $\tau$ (set to $0.6$ in all experiments) are dispatched to an external AI coding agent.
Experiments below $\tau$ are handled by a built-in multi-phase code agent that first emits a per-file blueprint, then generates files in dependency order using AST-derived summaries to maintain cross-file consistency.
Static validation gates check for detectable defects including identical ablation implementations and hardcoded metric values before any execution budget is spent.
A dedicated \emph{benchmark agent} handles dataset and baseline discovery; a \emph{figure agent} produces publication-quality visualizations.

\noindent \textbf{Sandboxed execution.}
All generated code runs in Docker containers under a three-phase network policy.
Phase~0 enables network access for dependency installation.
Phase~1 enables network access for data acquisition.
Phase~2 disables network access entirely during experiment execution, preventing both result exfiltration and pre-computed-result downloading.
Metric reporting is handled exclusively through a read-only evaluation harness, so generated code cannot redefine its own measurement infrastructure (Appendix~\ref{app:sandbox}).

\noindent \textbf{\textsc{Pivot}/\textsc{Refine} decisions.}
When an experiment fails or produces degenerate results, an automated repair loop captures the failure signature and generates targeted fixes.
The system then makes one of three decisions: \textsc{Proceed} when evidence supports the hypothesis, \textsc{Refine} when results are weak but the experimental direction is sound, or \textsc{Pivot} when the approach is fundamentally flawed, returning to hypothesis generation with the failure recorded as new evidence.
Systems that terminate on any failure avoid ambitious experiments by design.
By making failure recoverable, \system{} can pursue higher-risk hypotheses that would be abandoned under a brittle execution model.

\vspace{-0.3em}
\subsection{Verifiable Result Reporting}
\label{sec:verification}
\vspace{-0.3em}

LLM-generated papers face two integrity problems: fabricated experimental results and hallucinated citations.
Both arise from the same behavior---the model produces plausible-looking content with no grounding in actual evidence.
\system{} addresses both through deterministic verification gates applied at two granularities.

\noindent \textbf{Numeric registry.}
During execution, the system constructs a \emph{verified registry}: a whitelist of every value produced by experiment runs, storing per-condition means, standard deviations, and individual seed measurements.
At drafting time, pre-built \LaTeX{} tables populated exclusively from the registry are injected into the generation prompt.
After generation, a post-hoc verifier re-extracts every numeric claim and checks it against the registry, scoped per condition to prevent cross-condition false positives.
Claims in strict sections (Abstract, Results, Experiments) that cannot be matched to a registry entry trigger document rejection.
Claims in other sections are replaced with visible placeholders.
The writing agent can read the registry but cannot modify it.

\noindent \textbf{Citation verification.}
Every reference passes through a four-layer pipeline: DOI resolution via CrossRef, fuzzy title matching against OpenAlex, arXiv identifier lookup, and Semantic Scholar as a final fallback.
An LLM-based relevance check then classifies each reference as \textsc{Verified}, \textsc{Suspicious}, or \textsc{Hallucinated}.
References classified as \textsc{Hallucinated} are removed before any draft is finalized.

\vspace{-0.3em}
\subsection{Human-in-the-Loop Collaboration}
\label{sec:hitl}
\vspace{-0.3em}

Full automation reduces output quality at critical junctures where domain judgment matters.
Exhaustive step-by-step oversight eliminates the efficiency gains of automation.
The useful region lies between these extremes: human expertise is most valuable at a small number of high-leverage decision points rather than distributed uniformly across the pipeline.
\system{} provides seven intervention modes that let researchers select their operating point along this spectrum.

\noindent \textbf{Intervention modes.}
\emph{Full-Auto} runs the entire pipeline without human input.
\emph{Gate-Only} pauses at three fixed checkpoints: literature screening, experiment design, and final quality review.
\emph{Thorough} pauses at all phase boundaries, giving researchers visibility without requiring approval at every substep.
\emph{CoPilot} targets six high-leverage decision points, including hypothesis co-creation (\emph{Idea Workshop}), experiment design review (\emph{Baseline Navigator}), and collaborative paper drafting (\emph{Paper Co-Writer}).
\emph{Step-by-Step} requires explicit approval at every stage.
Two further regimes decompose CoPilot for the ablation in Section~\ref{sec:hitl_exp}: \emph{Pre-Experiment} retains intervention only at literature screening, hypothesis generation, and experiment design (early-pipeline), while \emph{Post-Experiment} retains intervention only at result analysis, paper draft, and quality gate (late-pipeline).
Our HITL ablation in Section~\ref{sec:hitl_exp} evaluates all seven modes empirically.

\noindent \textbf{SmartPause.}
Rather than relying on fixed checkpoints, SmartPause monitors the system's estimated uncertainty at each stage.
When uncertainty exceeds a learned threshold, the system pauses and presents the decision to the researcher.
The threshold adapts based on historical approval patterns: stages where the researcher frequently overrides the system are paused more often, while stages with consistently high approval rates proceed autonomously.

\vspace{-0.3em}
\subsection{Cross-Run Evolution}
\label{sec:evolution}
\vspace{-0.3em}

Existing autonomous research systems are stateless across runs: every run begins without knowledge of previous attempts, repeating failures that earlier runs already encountered.
\system{} maintains a persistent lesson store that converts past failures into future safeguards.

At the end of each run, the system extracts structured lessons from repair attempts, \textsc{Pivot}/\textsc{Refine} decisions, HITL gate feedback, and verification results.
Each lesson records a category, a severity score $s(l) \in (0, 1]$, and a recommended mitigation.
When a new run begins, relevant lessons are retrieved by category and ranked by a time-decayed weight:
\begin{equation}
w(l) \;=\; s(l) \cdot \exp\!\left(-\ln 2 \cdot \frac{\Delta t}{T_{1/2}}\right),
\label{eq:lesson-weight}
\end{equation}
where $\Delta t$ is the elapsed time since the lesson was recorded and $T_{1/2}$ is a half-life hyperparameter controlling how quickly older lessons lose influence (default $T_{1/2} = 30$ days).
Lessons are injected into prompts as natural-language overlays, requiring no model retraining and remaining applicable to any LLM backbone.
This design means that recent failures strongly constrain subsequent runs, while lessons from completed, successful lines of work gradually fade from prominence.

%% file: sections/experiment.tex
\vspace{-0.5em}
\section{Experiments}
\label{sec:experiments}
\vspace{-0.5em}

We evaluate \system{} through three complementary studies. First, we benchmark against existing systems on \bench{} using an experiment-stage evaluation, because most baselines cannot reliably produce complete papers without human supervision (\S\ref{sec:arcbench}). Second, we conduct an end-to-end evaluation from idea to paper on 10 \bench{} topics across seven human-in-the-loop regimes, assessing full paper quality under varying levels of intervention (\S\ref{sec:hitl_exp}). Third, we run a component ablation that isolates the contribution of each mechanism (\S\ref{sec:component_abl}). We close with a case study illustrating how the mechanisms interact on a single topic (\S\ref{sec:casestudy}).

\subsection{Experimental Setup}
\label{sec:setup}
\vspace{-0.3em}

\noindent \textbf{Benchmark.} We introduce \bench, a 25-topic ML benchmark (\textsc{ML01}--\textsc{ML25}) spanning tabular ML, optimization, dimensionality reduction, NLP, AutoML, GP kernels, topic modeling, semi-supervised learning, dynamical systems, anomaly detection, feature selection, causal discovery, and learning-to-rank, together with a 20-topic scientific-domain extension covering 10 high-energy physics (\textsc{P01}--\textsc{P10}), 7 systems biology (\textsc{B01}--\textsc{B07}), and 3 statistics (\textsc{S01}--\textsc{S03}) tasks. Each topic specifies a research question, a target dataset (or reference figure/simulation, for science topics), and expected experimental deliverables (code, results, analysis writeup). \bench{} supports three evaluation modes. The \emph{experiment-stage} mode evaluates systems at the experiment stage using a rubric-assisted strict judge, enabling fair comparison across systems with different end-to-end capabilities. The \emph{end-to-end} mode evaluates the full pipeline from research idea to completed paper, assessing overall paper quality on a 1--10 scale with accept rate ($\geq 5$) as the primary metric. The end-to-end mode is used both for the HITL ablation on 10 \textsc{ML} topics (\S\ref{sec:hitl_exp}) and the scientific-domain coverage study (\S\ref{sec:scidomain}). The \emph{scientific-domain} mode reuses the same code/exec/results/repro rubric on the 20 physics/biology/statistics topics, since most existing baselines cannot produce science-domain papers at all under fair-input conditions.

\noindent \textbf{Experiment-stage evaluation protocol.} The strict judge grades each (framework $\times$ topic) cell along three dimensions weighted as CD:CE:RA = 25:25:50. Code Development (CD) assesses whether the implementation correctly instantiates the proposed method and baselines. Code Execution (CE) verifies that experiments run to completion and produce valid outputs. Result Analysis (RA) evaluates whether conclusions are grounded in actual measurements, hypotheses receive explicit verdicts, and limitations are honestly reported. RA receives double weight because it captures the scientific reasoning quality that distinguishes autonomous research from automated scripting. Two independent agent reviewers run the strict judge in parallel; per-leaf disagreements exceeding $|\Delta| > 0.20$ are re-adjudicated and final scores are averaged. Details and inter-rater agreement are in Appendix~\ref{app:strict-judge}.

\noindent \textbf{Baselines and implementation.} We compare against AI Scientist v2~\citep{yamada2025aiscientistv2} and AIDE-ML~\citep{aide2025}, the two systems that provide end-to-end execution paths comparable to \system{} (Table~\ref{tab:comparison}). Agent Laboratory~\citep{schmidgall2025agentlab} is excluded because it does not deliver end-to-end execution under fair-input conditions. All frameworks use the same LLM backbone (GPT-5.3-codex) and the same sandboxed execution environment with identical per-experiment time budgets. This controlled setup isolates the contribution of system design from backbone capability.

\vspace{-0.3em}
\subsection{Main Results: Experiment-Stage Comparison}
\label{sec:arcbench}
\vspace{-0.3em}

Table~\ref{tab:arcbench-aggregate} reports mean scores across all 25 topics under the experiment-stage evaluation mode.

\begin{table}[t]
\centering
\caption{\bench{} experiment-stage results (25 topics, CD:CE:RA = 25:25:50).}
\label{tab:arcbench-aggregate}
\small
\begin{tabular*}{\textwidth}{@{\extracolsep{\fill}}lcccc@{}}
\toprule
\textbf{Framework} & \textbf{Code Dev} & \textbf{Code Exec}
  & \textbf{Result Analysis} & \textbf{Overall} \\
\midrule
\system{} (CoPilot)   & 0.968 & 0.578 & 0.523 & \textbf{0.648} \\
\system{} (Full-Auto) & 0.938 & 0.562 & 0.442 & 0.596 \\
AIDE-ML               & 0.958 & 0.415 & 0.336 & 0.511 \\
AI Scientist v2       & 0.712 & 0.442 & 0.261 & 0.419 \\
\bottomrule
\end{tabular*}
\end{table}

\begin{table}[t]
\centering
\caption{End-to-end HITL ablation across 10 topics and 7 intervention regimes.
Paper quality scored 1--10; accept $=$ score $\geq 5$.}
\label{tab:hitl-summary}
\small
\begin{tabularx}{\textwidth}{@{}lCCCC@{}}
\toprule
\textbf{Mode} & \textbf{Valid} & \textbf{Mean Q} & \textbf{Accept} & \textbf{Interventions} \\
\midrule
Full-Auto        & 8/10  & 4.03          & 25.0\%          & 0  \\
Gate-Only        & 10/10 & 5.03          & 50.0\%          & 3  \\
\textbf{CoPilot} & 8/10  & \textbf{7.27} & \textbf{87.5\%} & 6 \\
Thorough         & 7/10  & 4.86          & 42.9\%          & 8 \\
Step-by-Step     & 10/10 & 5.19          & 50.0\%          & 23 \\
Pre-Experiment   & 8/10  & 4.28          & 37.5\%          & 3 \\
Post-Experiment  & 6/10  & 5.08          & 50.0\%          & 3 \\
\bottomrule
\end{tabularx}
\end{table}

\noindent \textbf{\system{} outperforms all baselines across all dimensions.} \system{} (CoPilot) achieves the highest overall strict score (0.648), outperforming AI Scientist v2 (0.419) by 54.7\% and AIDE-ML (0.511) by 26.8\%. Even in Full-Auto mode without human intervention, \system{} (0.596) substantially exceeds both baselines, indicating that the gains are primarily driven by system design rather than human input.

\noindent \textbf{The largest advantage is on Result Analysis.} The performance gap is most pronounced on Result Analysis, where \system{} (CoPilot) scores 0.523 against AI Scientist v2's 0.261, a 100.4\% relative improvement. This dimension evaluates whether conclusions are hypothesis-aligned, tables contain only verified numbers, and limitations are honestly reported. The advantage directly reflects multi-agent debate at the result analysis stage and the verified result registry: debate forces explicit per-hypothesis verdicts with critical scrutiny, and the registry ensures every reported number traces to an actual measurement. In contrast, AI Scientist v2's single-agent analysis tends to oversell weak findings without cross-examination.

\noindent \textbf{Code Development is competitive; execution separates the systems.} All systems score above 0.70 on Code Development, with AIDE-ML (0.958) nearly matching \system{} (0.968). The differentiator is what happens after code is written. AIDE-ML's execution success rate (0.415) is substantially lower, reflecting its lack of self-healing: experiments that encounter runtime errors are discarded rather than repaired. \system{}'s self-healing executor raises execution success to 0.562 (Full-Auto) and 0.578 (CoPilot) by diagnosing failures and applying targeted fixes through the \textsc{Pivot}/\textsc{Refine} loop.

\noindent \textbf{Failure mode analysis.} Among the 25 topics, \system{} (Full-Auto) fails to produce valid results on 2 topics, both involving complex multi-file implementations with cascading dependencies. AI Scientist v2 fails on 6 topics, with failures concentrated in topics requiring iterative experiment refinement (dynamical systems, causal discovery) where single-attempt execution without recovery is insufficient. This pattern confirms that self-healing is most valuable on topics where the first implementation attempt is unlikely to succeed.

\vspace{-0.3em}
\subsection{Cross-Domain Coverage: Physics, Biology, and Statistics}
\label{sec:scidomain}
\vspace{-0.3em}

The \bench{} core (\textsc{ML01}--\textsc{ML25}) is intentionally ML-focused to enable a fair comparison against AIDE-ML and AI Scientist v2, both of which target ML pipelines.
Autonomous research systems must, however, operate across scientific domains.
We therefore extend \bench{} with 20 \emph{scientific-domain} tasks that each require domain-specific software stacks: \textsc{MadGraph5\_aMC@NLO}~\citep{Alwall2014} / \textsc{Pythia8} / \textsc{Delphes} / \textsc{FeynRules} / \textsc{MadAnalysis5} for high-energy physics; \textsc{COBRApy}~\citep{Ebrahim2013} / \textsc{BiGG}~\citep{King2016} genome-scale models / \textsc{optlang} / GLPK for systems biology; and double-machine-learning~\citep{Chernozhukov2018}, bootstrap~\citep{Efron1979}, and selective-inference machinery for statistics.

\system{} routes each topic through a \emph{sandboxed, domain-specialized agent} during the experiment stage.
The HEP agent is equipped with the FeynRules, MadGraph, and MadAnalysis skills drawn from \citet{qiu2026endtoendarchitecturecolliderphysics}, which introduced the ColliderAgent architecture we directly adopt here; the biology agent is equipped with GEM-builder, flux-balance analysis, and flux-analyzer skills; the statistics agent is equipped with Monte Carlo simulation and semiparametric inference skills.
Each specialized agent runs inside a Claude Code subprocess with the requisite packages pre-installed, so the top-level orchestrator requires no domain-specific knowledge.
This design is what enables \system{} to reproduce experiments across heterogeneous scientific fields without per-domain engineering effort.

Each task is graded with the same code / execution / results / repro rubric used in Table~\ref{tab:arcbench-aggregate}, with leaf criteria rewritten to reflect domain-appropriate targets (e.g., ``loaded the correct \textsc{iJO1366} model and set anaerobic medium bounds'' for biology; ``computed cross-section ratios within 3\% of the published reference'' for HEP).
Per-task scores are aggregated to a column mean per domain.
Table~\ref{tab:scidomain} reports the results.
AIDE-ML and AI Scientist v2 fail to install the required HEP and biology stacks under fair-input conditions and produce no valid output on \textsc{P01}--\textsc{P10} and \textsc{B01}--\textsc{B07}; on the statistics tasks both systems execute but their generic ML scaffolding yields incomplete implementations that miss the inferential targets (e.g., bias and coverage metrics not computed, no Neyman-orthogonal score, no estimand writeup).

\begin{table}[t]
\centering
\caption{Scientific-domain coverage. Scores are computed with the same
code:exec:results:repro rubric as Table~\ref{tab:arcbench-aggregate}.
We evaluate biology on \textsc{B01}--\textsc{B07}, statistics on
\textsc{S01}--\textsc{S03}, and HEP-ph on \textsc{P01}--\textsc{P10}
. A red cross indicates a code-execution failure caused by
missing or unusable domain-specific software.
Statistics tasks do not require specialized simulation stacks, so all systems
are scored normally.
Failed runs are counted as zero when computing the run-weighted overall mean
over the 20 science-domain tasks.}
\label{tab:scidomain}
\small
\begin{tabular*}{\textwidth}{@{\extracolsep{\fill}}lcccc@{}}
\toprule
\textbf{Framework} & \textbf{Biology} & \textbf{Statistics}
  & \textbf{HEP-ph} & \textbf{Overall} \\
\midrule
\system{} (CoPilot) & \textbf{0.912} & \textbf{0.898} & \textbf{0.489} & \textbf{0.867} \\
AIDE-ML             & \xmark        & 0.452          & \xmark        & 0.090 \\
AI Scientist v2     & \xmark        & 0.418          & \xmark        & 0.084 \\
\bottomrule
\end{tabular*}
\end{table}

\noindent \textbf{Sandboxed domain agents are necessary for cross-domain coverage.}
The biology column (mean 0.912 ranging from \emph{E.~coli} succinate knockout screens to \emph{M.~tuberculosis} essentiality and drug-target prioritization) is supported entirely by COBRApy-skilled sub-agents operating over BiGG genome-scale models; the orchestrator is never required to manipulate stoichiometric matrices directly.
The statistics column (mean 0.898 on bootstrap confidence-interval coverage and DML/AIPW estimation) is supported by the Monte Carlo and semiparametric-inference skill bundle; the third statistics run completed the pipeline but failed the requirements judge on missing metric artifacts and is therefore excluded from the column mean.
The HEP column requires Lagrangian-to-UFO translation, MadGraph parton-level generatio, and quantitative reproduction of a published cross-section curve; \system{} correctly reproduces the predicted shape and numerical cross-section values, but incurs scoring penalties for insufficient deliverable content and minor unsupported meta-claims.
The qualitative conclusion of Table~\ref{tab:scidomain} is that both baselines score zero on physics and biology because their sandboxes do not include the required scientific software; \system{}'s domain-skill installation step closes this gap.

\vspace{-0.3em}
\subsection{End-to-End HITL Ablation}
\label{sec:hitl_exp}
\vspace{-0.3em}

\bench{}'s experiment-stage mode enables fair cross-system comparison but does not evaluate the full research pipeline. To assess end-to-end quality from idea to completed paper, we run a HITL ablation on 10 \bench{} topics across seven intervention regimes, ranging from \textbf{Full-Auto} (zero interventions) to \textbf{Step-by-Step} (every stage). \textbf{CoPilot} targets six high-leverage decision points; \textbf{Pre-Experiment} and \textbf{Post-Experiment} isolate early-stage and late-stage contributions respectively. Each mode receives the same payload at its covered stages. Setup details are in Appendix~\ref{app:hitl-setup}.

\noindent \textbf{More intervention does not monotonically improve quality.} Table~\ref{tab:hitl-summary} reports mode-level summaries. CoPilot achieves the highest mean paper-quality score (7.27) and accept rate (87.5\%) with 19 targeted interventions. Step-by-Step requires 29 interventions but achieves only 5.19 and 50\%. On matched topics, CoPilot beats Full-Auto by $+3.21$ and beats Step-by-Step by $+2.16$. The explanation is that Step-by-Step's approve actions at non-critical stages add noise without information, while CoPilot concentrates expert judgment where it has the highest marginal impact.

\noindent \textbf{Pre-Experiment and Post-Experiment HITL address different failure modes.} Splitting CoPilot reveals two complementary contributions. Pre-Experiment HITL (stages 5, 8, 9) fixes research design feasibility: on T02, it compresses a 240-cell factorial design to a feasible 60-cell layout and prescribes appropriate statistical tests. Pre-Experiment alone is widely valid (8/10) but rarely lifts quality (mean 4.28, 37.5\% accept), because correct scoping alone does not fix downstream claim discipline. Post-Experiment HITL (stages 14, 17, 20) fixes claim faithfulness, ensuring conclusions match actual measurements and scope is honestly delimited. Post-Experiment produces some of the strongest individual papers but is valid on only 6/10 topics, because late-stage HITL cannot generate experimental evidence from nothing. CoPilot is best because it spans both halves: fixing feasibility early and faithfulness late.

\noindent \textbf{Gate-Only provides a cost-effective middle ground.} Gate-Only (3 interventions) raises accept rate from 25\% to 50\% and is the only mode achieving 10/10 validity. For practitioners seeking minimal human involvement with meaningful quality improvement, Gate-Only represents an attractive operating point between Full-Auto and CoPilot.

\vspace{-0.3em}
\subsection{Component Ablation}
\label{sec:component_abl}
\vspace{-0.3em}

To isolate the contribution of each mechanism, we run a system-level ablation on the same 10 ARC-BENCH topics under Full-Auto mode. Each row in Table~\ref{tab:component-abl} removes one mechanism and keeps the others intact.

\noindent \textbf{Best-of-$N$ protocol.}
Autonomous research agents are stochastic and path-dependent: a single unlucky branch in hypothesis generation, code repair, or drafting can dominate the final outcome. We therefore adopt a best-of-$N$ protocol; Full-Auto numbers in this ablation reflect this setting and should be read against Table~\ref{tab:component-abl} rather than the single-run HITL results. Detailed ablations of design-space exploration parameters appear in Appendix~\ref{app:agent_count}.

\begin{table}[t]
\centering
\caption{Component ablation in Full-Auto mode on the same 10 ARC-BENCH topics using a best-of-3 protocol over three reruns per configuration--topic pair. Completion counts topics for which at least one rerun completed a manuscript. Quality is the mean of the selected best scores over completed topics, and Accept counts selected completed outputs with score $\ge 5$. Fabrication is determined by manual audit of the selected outputs. $^{\ddagger}$~Score inflated by removing the verification gate.}
\label{tab:component-abl}
\small
\begin{tabularx}{\textwidth}{@{}lCCCC@{}}
\toprule
\textbf{Configuration} & \textbf{Completion} & \textbf{Quality} & \textbf{Accept} & \textbf{Fabrication} \\
\midrule
Full \system{}              & \textbf{10/10} & \textbf{5.62} & \textbf{3/10}             & \xmark \\
\quad w/o Debate            & 10/10          & 4.25          & 1/10                      & \xmark \\
\quad w/o Self-Healing      & 6/10           & 4.83          & 1/6                       & \xmark \\
\quad w/o Evolution         & 9/10           & 5.14          & 2/10                      & \xmark \\
\quad w/o Verification      & 10/10          & $5.48^{\ddagger}$ & $5/10^{\ddagger}$     & \cmark \\
\quad w/o Debate \& Healing & 4/10           & 3.47          & 0/4                       & \xmark \\
\bottomrule
\end{tabularx}
\end{table}

\noindent \textbf{Each mechanism addresses a distinct failure mode, even under a favorable rerun budget.}
Multi-agent debate is the largest quality contributor ($-1.37$, $p{=}0.003$): without the Pragmatist filtering infeasible hypotheses and the Skeptic scrutinizing weak findings, both hypothesis quality and result analysis degrade. Self-healing is the largest completion contributor: removing it reduces completion from 10/10 to 6/10 even with three attempts per topic, because the first unrecovered runtime error terminates the run and surviving topics are disproportionately easier. Cross-run evolution provides a moderate reliability gain ($-0.48$ quality, $-1$ completion) by injecting lessons from prior failures, primarily avoiding known failure modes rather than raising the quality ceiling.

\noindent \textbf{Verification is the integrity backstop.}
Removing the verified registry raises apparent acceptance from 3/10 to 5/10, but manual inspection reveals that 3 of those 5 papers contain values absent from any measurement record. The verification gate correctly separates genuine from fabricated results; its cost to acceptance rate is the price of scientific integrity.

\noindent \textbf{The mechanisms interact super-additively.}
Combined removal of debate and self-healing drops completion to 4/10, mean quality to 3.47, and acceptance to zero. The interaction is intuitive: debate without self-healing produces ambitious hypotheses that crash on first failure, while self-healing without debate repairs experiments that test poorly formulated questions.

\vspace{-0.3em}
\subsection{Case Study: Topic T10}
\label{sec:casestudy}
\vspace{-0.3em}

\begin{figure}
    \centering
    \includegraphics[width=0.9\linewidth]{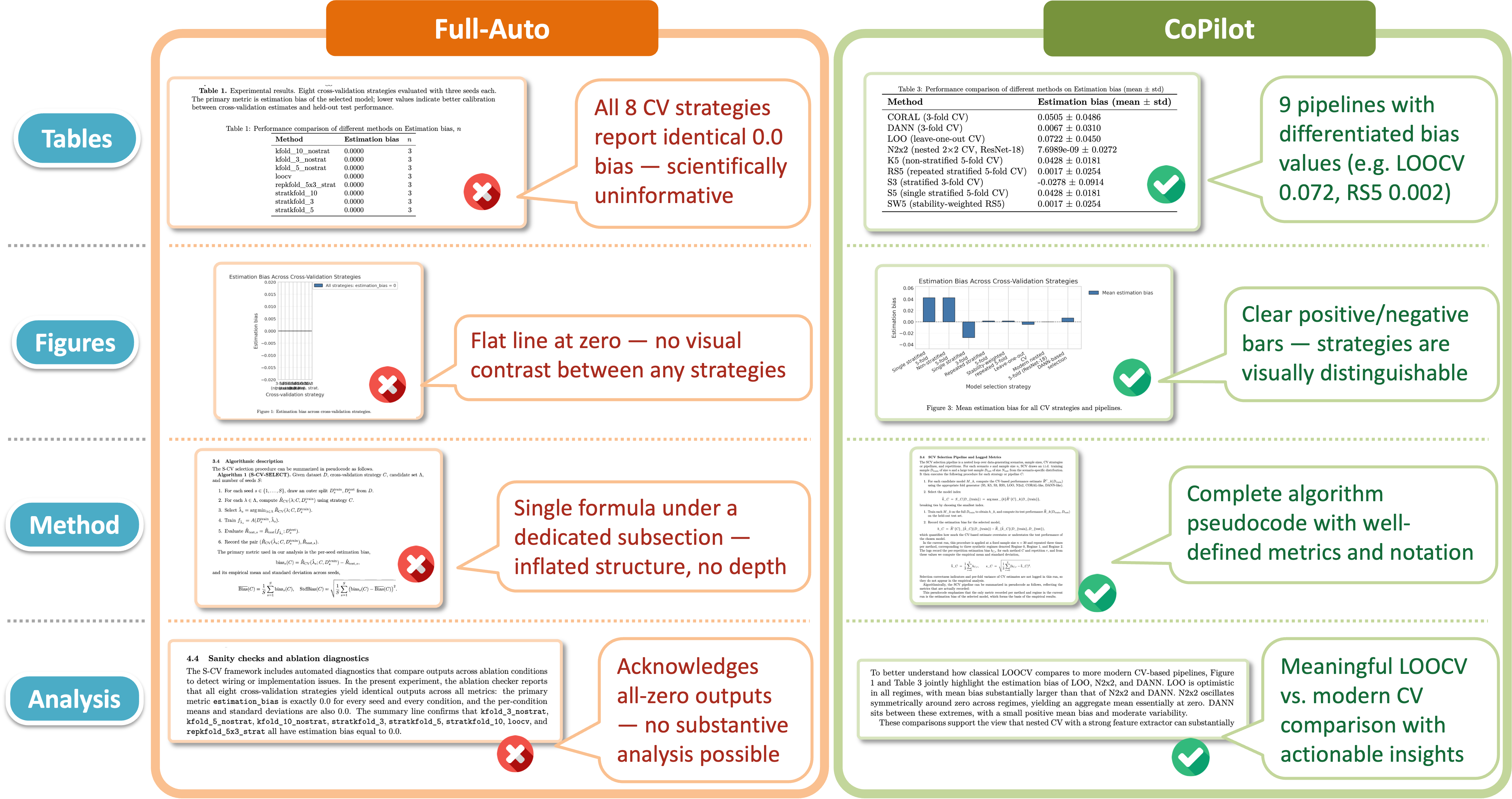}
    \caption{Case study comparing Full-Auto and CoPilot on Topic T10. Full-Auto suffers a silent semantic collapse: all cross-validation strategies report identical zero-bias outputs. CoPilot produces differentiated results, enabling a meaningful comparison across strategies.}
    \label{fig:case-study}
\end{figure}

Figure~\ref{fig:case-study} compares Full-Auto and CoPilot on Topic T10, which studies cross-validation strategies for small-sample model selection. Both runs produce complete-looking manuscripts, but Full-Auto fails in a way that runtime metrics do not capture: its experiment collapses to identical outputs across conditions. A detailed artifact-level comparison is provided in Appendix~\ref{app:casestudy}.

The T10 case shows why execution success alone is not enough. Full-Auto completes a manuscript, but all eight cross-validation strategies collapse to identical zero-bias outputs. As a result, the paper cannot support a substantive comparison. CoPilot avoids this failure because the human guidance targets the experimental bottleneck directly. It asks the system to check whether cross-validation strategies produce different outcomes, whether leave-one-out cross-validation fits the time budget, and whether the manuscript's claims stay within the logged results. The final CoPilot paper remains exploratory, but it reports nonzero contrasts across nine pipelines and states its limitations clearly. The Stage-20 quality gate therefore accepts it.

Three observations follow. First, debate quality matters even when execution succeeds: the Contrarian's concern about distinguishable ablations helps flag the collapsed design before it becomes the final experiment. Second, verification is necessary but not sufficient. Full-Auto passes the numeric gate because the zero values are real logged measurements, not fabricated numbers. However, the gate cannot tell whether those measurements answer the research question. Third, CoPilot improves quality not by adding more intervention, but by placing intervention at the right decision points. Guidance about experimental semantics produces evidence-grounded output, while late writing-stage guidance cannot recover missing experimental evidence.

This pattern is consistent with the HITL results in Table~\ref{tab:hitl-summary}: partial or mistimed intervention improves some failure modes, but does not match CoPilot's combination of early experimental guidance and late claim checking. Broader failure and writing-quality audits are reported in Appendices~\ref{app:failures} and~\ref{app:writing-quality}.

%% file: sections/analysis.tex

%% file: sections/conclusion.tex
\vspace{-0.5em}
\section{Conclusion}
\label{sec:conclusion}
\vspace{-0.5em}

We presented \system{}, a multi-agent autonomous research pipeline that unifies structured debate, self-healing execution, verifiable result reporting, cross-run evolution, and human-in-the-loop collaboration in a single self-reinforcing system. On \bench, \system{} outperforms AI Scientist v2 by 54.7\%, with the largest gains on result analysis where multi-agent debate and verified reporting produce hypothesis-aligned, grounded conclusions. An end-to-end HITL ablation across seven intervention regimes shows that targeted intervention at high-leverage decision points (CoPilot, 87.5\% accept rate) consistently outperforms both full autonomy (25\%) and exhaustive step-by-step oversight (50\%), establishing that precise human-AI collaboration is a more effective paradigm than either extreme. Component ablation confirms that the mechanisms are complementary: debate drives quality, self-healing drives completion, verification enforces integrity, and their combined removal is super-additive. We position \system{} as a research amplifier that accelerates scientific exploration while keeping verifiability at the center, rather than a replacement for human scientific judgment. Detailed ethical issues are discussed in Appendix~\ref{sec:ethics}.

%% file: sections/appendix.tex

\section{Full Stage Definitions}
\label{app:stages}

Table~\ref{tab:all_stages} provides the complete specification of all 23 pipeline
stages. The 23-stage design reflects a tension between granularity and overhead.
Early prototypes used a coarser 12-stage pipeline, but bundling too many
responsibilities into a single LLM call (\eg{} ``search the literature, screen it,
and extract knowledge'' as one stage) led to poor intermediate quality that
compounded downstream. Conversely, very fine-grained pipelines (we experimented
with 30+) incurred excessive context-reconstruction overhead. The current 23-stage
design was arrived at iteratively. Each stage defines a formal \emph{contract}: a
JSON schema specifying required input fields (with types and validation), expected
output fields, acceptance criteria (\eg{} ``at least 2 hypotheses must be marked
falsifiable''), and an error-code namespace (\eg{} \texttt{E-HYPO-*}).

\begin{table}[h]
\centering
\caption{Complete 23-stage pipeline specification.}
\label{tab:all_stages}
\small
\begin{tabular}{@{}c l l l@{}}
\toprule
\textbf{\#} & \textbf{Stage Name} & \textbf{Key Output} & \textbf{Special} \\
\midrule
1 & \textsc{Topic\_Init} & SMART goal, hardware profile & Domain detect \\
2 & \textsc{Problem\_Decompose} & Problem tree & \\
3 & \textsc{Search\_Strategy} & Search queries, sources.json & \\
4 & \textsc{Literature\_Collect} & candidates.jsonl & API calls \\
5 & \textsc{Literature\_Screen} & Shortlisted papers & HITL Gate \\
6 & \textsc{Knowledge\_Extract} & Knowledge cards & \\
7 & \textsc{Synthesis} & Gap analysis & \\
8 & \textsc{Hypothesis\_Gen} & 2--4 hypotheses & Debate ($K{=}3$) \\
9 & \textsc{Experiment\_Design} & exp\_plan.yaml + benchmarks & HITL Gate \\
10 & \textsc{Code\_Generation} & Multi-file experiment code & Cascade$^\dagger$ \\
11 & \textsc{Resource\_Planning} & Time/resource estimates & \\
12 & \textsc{Experiment\_Run} & results.json per condition & Docker sandbox \\
13 & \textsc{Iterative\_Refine} & Improved code + metrics & Self-heal \\
14 & \textsc{Result\_Analysis} & analysis.md + charts & Debate ($K{=}3$) \\
15 & \textsc{Research\_Decision} & \textsc{Proceed}/\textsc{Refine}/\textsc{Pivot} & Decision \\
16 & \textsc{Paper\_Outline} & Section outline & \\
17 & \textsc{Paper\_Draft} & Full paper draft & Verified tables \\
18 & \textsc{Peer\_Review} & Review feedback & Multi-agent \\
19 & \textsc{Paper\_Revision} & Revised paper & Length guard \\
20 & \textsc{Quality\_Gate} & quality\_report.json & HITL Gate \\
21 & \textsc{Knowledge\_Archive} & Lessons extracted & Noncritical \\
22 & \textsc{Export\_Publish} & \LaTeX{} + sanitized BibTeX & Anti-fabrication \\
23 & \textsc{Citation\_Verify} & verification\_report.json & 4-layer API \\
\bottomrule
\end{tabular}
\vspace{0.3em}

\noindent{\footnotesize $^\dagger$Cascade: Beast Mode (external AI agent) $\to$ CodeAgent (multi-phase,
blueprint, sequential, AST gates) $\to$ Legacy single-shot. Each tier falls back
to the next on failure.}
\end{table}

\begin{algorithm}[t]
\caption{\system{}: Autonomous Research Pipeline with Domain-Aware Routing}
\label{alg:pipeline}
\small
\begin{algorithmic}[1]
\REQUIRE Research idea $\mathcal{I}$, lesson store $\mathcal{L}$, max pivots $N_p{=}2$, max refines $N_r{=}10$
\ENSURE Verified paper $\mathcal{P}$, updated lesson store $\mathcal{L}'$
\STATE Detect hardware profile; detect research domain $D$ via three-level cascade
       (forced override $\to$ keyword matching $\to$ LLM classification)
\STATE Select domain prompt bank $\mathcal{B}(D)$; instantiate \textsc{PromptManager}$(D)$
       and domain adapter $\mathcal{A}(D)$; inject relevant lessons from $\mathcal{L}$
\STATE Scope topic from $\mathcal{I}$ using $\mathcal{B}(D)$.\textsc{topic\_init};
       retrieve \& screen literature; synthesise knowledge cards
\STATE $n_p \gets 0$
\REPEAT
    \STATE \textbf{[Debate]} $K{=}3$ domain-specialised agents via $\mathcal{A}(D)$.\texttt{debate\_roles\_hypothesis}
           $\to$ hypotheses $\mathcal{H}$ (e.g.\ Innovator/Pragmatist/Contrarian for ML;
           Theorist/Phenomenologist/Experimentalist for HEP-ph)
    \STATE Design experiments from $\mathcal{H}$ with $\mathcal{A}(D)$-injected condition
           terminology, paradigm constraints, and compute budget
    \STATE Score complexity $c$; select generator via cascade (Beast\,$\to$\,Agent\,$\to$\,Legacy)
    \STATE Validate code: AST checks, import verification, ablation-identity detection
    \STATE $n_r \gets 0$
    \REPEAT
        \STATE Execute in Docker sandbox (three-phase network isolation) with
               domain-specific image and pre-cached datasets
        \WHILE{experiment fails \AND repair budget remains}
            \STATE Parse failure signature; generate targeted fix; re-execute
        \ENDWHILE
        \STATE \textbf{[Debate]} $K{=}3$ agents via $\mathcal{A}(D)$.\texttt{debate\_roles\_analysis}
               $\to$ result analysis
        \STATE Decide $d \in \{\textsc{Proceed}, \textsc{Refine}, \textsc{Pivot}\}$; $n_r \gets n_r + 1$
    \UNTIL{$d \neq \textsc{Refine}$ \OR $n_r \geq N_r$}
    \STATE $n_p \gets n_p + 1$
\UNTIL{$d \neq \textsc{Pivot}$ \OR $n_p \geq N_p$}
\STATE Build verified registry $\mathcal{R}$ from experiment results
\STATE Draft paper with pre-built tables from $\mathcal{R}$ using $\mathcal{A}(D)$.\texttt{preferred\_template};
       review $\to$ revise
\STATE Verify: reject if unverified numbers appear in strict sections
\STATE Verify citations via 4-layer API pipeline (CrossRef, OpenAlex, arXiv, S2)
\STATE Extract lessons; update $\mathcal{L} \to \mathcal{L}'$ with time-decayed weights
\RETURN $\mathcal{P}$ with verification reports
\end{algorithmic}
\end{algorithm}



\section{Prompt Design}
\label{app:prompts}

\subsection{Prompt Architecture}
\label{app:prompt-arch}

The prompt system is organised into three nested layers.

\paragraph{Primary layer.}
Twenty-three stage-specific prompts, each comprising a system message, a structured user template, an optional \texttt{json\_mode} flag, and a \texttt{max\_tokens} override.
Section-level word-count targets are enforced post-generation: abstract 150--200, introduction 800--1000, related work 600--800, method 1000--1500, experiments 800--1200, results 600--800, discussion 400--600, conclusion 200--300.

\paragraph{Reusable block layer.}
Shared text fragments injected into multiple stages to enforce consistent writing quality and experimental rigour: \texttt{academic\_style\_guide}, \texttt{anti\_hedging\_rules}, \texttt{writing\_structure}, \texttt{dataset\_guidance}, and \texttt{hp\_reporting}.

\paragraph{Sub-prompt layer.}
Specialised prompts for intra-stage operations: \texttt{architecture\_planning}, \texttt{generate\_single\_file}, \texttt{code\_repair}, \texttt{hypothesis\_synthesize}, \texttt{analysis\_synthesize}, and \texttt{iterative\_improve}.
Total prompt text across all three layers is approximately 46K tokens.
Users can override any prompt via \texttt{prompts.default.yaml}; the \textsc{PromptManager} safely renders \texttt{\{word\_chars\}} placeholders while leaving JSON schema syntax untouched.

\paragraph{Domain-aware prompt selection.}
The \textsc{PromptManager} accepts a \texttt{domain} argument at instantiation and selects the corresponding stage bank.
Two native banks exist: an ML bank (default) and a HEP-ph phenomenology bank.
All other domains receive the ML bank augmented with a domain-adapter overlay.
Both banks expose the same 23 stage keys and the same \texttt{\{placeholder\}} variables, enforced by a parity test suite, so all downstream pipeline logic is bank-agnostic.

Domain detection runs a three-level cascade at pipeline start.
\emph{Level~0}: a forced override via \texttt{project.profile} in \texttt{config.yaml} skips all detection and ensures cross-stage consistency.
\emph{Level~1}: fast keyword matching against 350+ domain-specific rules ordered most-specific-first.
\emph{Level~2}: LLM classification for topics not resolved by keywords, outputting one of 24 domain IDs.
\emph{Level~3}: generic \texttt{\_generic.yaml} fallback.

Beyond the two native banks, \system{} supports 20+ domains through a profile-and-adapter system in \texttt{researchclaw/domains/}.
Each domain profile (a YAML file) specifies the experiment paradigm, condition terminology, standard baselines, typical file structure, core libraries, Docker image, metric types, statistical tests, output formats, and prompt guidance blocks.
Each domain adapter injects domain-specific content into stage prompts via \texttt{\{domain\_context\}} and related placeholders.
Table~\ref{tab:domain-support} lists domains with complete end-to-end configuration.

\begin{table}[h]
\centering
\caption{Domain support matrix. \emph{Bank}: native 23-stage prompt bank. \emph{Profiles}: YAML profiles with experiment-paradigm configuration. \emph{Adapter}: Python adapter with stage-level prompt injection. \emph{Template}: preferred journal \LaTeX{} template.}
\label{tab:domain-support}
\small
\begin{tabular}{@{}lllll@{}}
\toprule
\textbf{Domain} & \textbf{Bank} & \textbf{Profiles} & \textbf{Adapter} & \textbf{Template} \\
\midrule
Machine Learning        & ML            & 8  & \checkmark & NeurIPS/ICML/ICLR \\
High-Energy Physics     & HEP-ph        & 1  & \checkmark & JHEP / PRD \\
Computational Biology   & Bio  & 3  & \checkmark & Bioinformatics \\
Computational Chemistry & Chem  & 2  & \checkmark & JCTC / JCIM \\
Theoretical Physics     & Phy  & 1  & \checkmark & PRX / J.\ Comp.\ Phys. \\
Empirical Economics     & ML + adapter  & 1  & \checkmark & AER / JEEA \\
Mathematics             & ML + adapter  & 2  & \checkmark & Math.\ Comp. \\
Computational Neuroscience & ML + adapter & 1 & \checkmark & PLoS Comp.\ Bio. \\
\bottomrule
\end{tabular}
\end{table}

\paragraph{Domain-differentiated debate roles.}
The most consequential domain-level difference is in multi-agent debate personas.
For hypothesis generation (Stage~8), the ML bank uses \emph{Innovator} (cross-domain analogies, high-risk hypotheses), \emph{Pragmatist} (computational feasibility, incremental gains), and \emph{Contrarian} (challenges assumptions, finds failure modes).
The HEP-ph bank replaces these with \emph{Theorist} (BSM Lagrangians, symmetries, UV completions), \emph{Phenomenologist} (testable observable signatures, analytical cross-section estimates), and \emph{Experimentalist} (detector acceptances, background shapes, overlooked systematics).
For result analysis (Stage~14), ML uses Optimist / Skeptic / Methodologist; HEP-ph uses Model-Builder / Phenomenologist / Experimentalist, with all verdicts stated in natural units with explicit experimental bound citations.

\subsection{Per-Stage Prompt Breakdown}
\label{app:prompt-stages}

Table~\ref{tab:ml-prompts} lists all 23 ML-bank stages with their system role, key prompt requirements, and output format.
The full bank is approximately 28K tokens.
Two stages with the highest prompt engineering load are detailed below the table.

\begin{table}[htbp]
\centering
\caption{ML prompt bank: per-stage system role, key user requirements, and output format.}
\label{tab:ml-prompts}
\small
\begin{tabular}{@{}clp{5.5cm}l@{}}
\toprule
\textbf{\#} & \textbf{Stage} & \textbf{Key requirements} & \textbf{Output} \\
\midrule
\multicolumn{4}{@{}l}{\textit{Phase A: Research Scoping}} \\
1 & \textsc{Topic\_Init}
  & Novel angle, SMART goal, trend validation, benchmark specification
  & Markdown \\
2 & \textsc{Problem\_Decompose}
  & $\geq$4 prioritised sub-questions, risks
  & Markdown \\
\midrule
\multicolumn{4}{@{}l}{\textit{Phase B: Literature Discovery}} \\
3 & \textsc{Search\_Strategy}
  & Merged search plan, source manifest
  & JSON \\
4 & \textsc{Literature\_Collect}
  & $\geq$8 candidate papers with abstract, year, source
  & JSON \\
5 & \textsc{Literature\_Screen}
  & Domain-match filter, recency preference, quality floor
  & JSON \\
6 & \textsc{Knowledge\_Extract}
  & Structured knowledge cards: problem, method, data, metrics, findings
  & JSON \\
\midrule
\multicolumn{4}{@{}l}{\textit{Phase C: Knowledge Synthesis}} \\
7 & \textsc{Synthesis}
  & Cluster overview, gap list, prioritised opportunities
  & Markdown \\
8 & \textsc{Hypothesis\_Gen}
  & $\geq$2 falsifiable hypotheses; novelty argument, measurable prediction, failure condition, required baselines
  & Markdown \\
\midrule
\multicolumn{4}{@{}l}{\textit{Phase D: Experiment Design}} \\
9 & \textsc{Experiment\_Design}
  & YAML plan: objectives, datasets, baselines, proposed methods, ablations, metrics, risks, compute budget
  & YAML \\
10 & \textsc{Code\_Generation}
  & Multi-file Python; algorithm integrity; calibration loop; PyTorch detach rules; condition breadth-first ordering
  & Multi-file code \\
11 & \textsc{Resource\_Planning}
  & GPU/CPU time estimates, parallelism plan, fallback strategy
  & JSON \\
\midrule
\multicolumn{4}{@{}l}{\textit{Phase E: Experiment Execution}} \\
12 & \textsc{Experiment\_Run} & Execution stage; no LLM call & \emph{metrics.json} \\
13 & \textsc{Iterative\_Refine}
  & Targeted fix based on failure signature; repair budget tracking
  & Patched code \\
\midrule
\multicolumn{4}{@{}l}{\textit{Phase F: Analysis \& Decision}} \\
14 & \textsc{Result\_Analysis}
  & Per-hypothesis verdict; effect sizes; methodology audit; quality rating 1--10; $K{=}3$ debate consensus
  & Markdown \\
15 & \textsc{Research\_Decision}
  & \textsc{Proceed} / \textsc{Refine} / \textsc{Pivot} with evidence-based justification
  & Markdown \\
\midrule
\multicolumn{4}{@{}l}{\textit{Phase G: Paper Writing}} \\
16 & \textsc{Paper\_Outline}
  & Section structure, 3 title candidates, per-section content plan
  & Markdown \\
17 & \textsc{Paper\_Draft}
  & Full 8--10 page draft; verified-registry tables only; no fabricated numbers in strict sections
  & Markdown \\
18 & \textsc{Peer\_Review}
  & $\geq$3 reviewer perspectives; soundness, novelty, presentation, significance
  & Markdown \\
19 & \textsc{Paper\_Revision}
  & Point-by-point response to reviews; section-length guard
  & Markdown \\
\midrule
\multicolumn{4}{@{}l}{\textit{Phase H: Finalization}} \\
20 & \textsc{Quality\_Gate} & Quality report; noncritical warnings do not block & JSON \\
21 & \textsc{Knowledge\_Archive} & Lesson extraction; time-decayed update & Markdown \\
22 & \textsc{Export\_Publish}
  & \LaTeX{} + sanitised BibTeX; domain-template selection; anti-fabrication pass
  & \LaTeX{} \\
23 & \textsc{Citation\_Verify}
  & 4-layer API pipeline; reject unresolvable DOI in strict sections
  & JSON \\
\bottomrule
\end{tabular}
\end{table}

\begin{tcolorbox}[
  title={Stage~1 Deep Dive: \textsc{Topic\_Init}},
  colback=blue!3!white,
  colframe=blue!50!black,
  fonttitle=\bfseries\small,
  breakable,
  label={box:stage1-prompt},
]
Stage~1 illustrates the prompt structure shared across all stages.
The \emph{system} message defines the agent role and novelty principles:

\smallskip
\begin{quote}\small\itshape
``You are a rigorous research planner who identifies NOVEL, TIMELY research angles.
You follow recent trends from top venues in the relevant domain and propose research that advances the frontier rather than repeating known results.
NOVELTY PRINCIPLES: A good research angle addresses a GAP not yet covered by existing work.
Avoid pure benchmark/comparison studies unless the methodology is novel.
The research must be FEASIBLE with limited compute (single GPU, hours not days).
Check: would a reviewer say `this is already well-known'? If so, find a sharper angle.''
\end{quote}
\smallskip

The \emph{user} template is parameterised by \texttt{\{topic\}}, \texttt{\{domains\}}, \texttt{\{project\_name\}}, and \texttt{\{quality\_threshold\}}, and requires six output sections:
\textbf{Topic}, \textbf{Novel Angle} (specific unexplored aspect with trend validation---no fabricated paper titles), \textbf{Scope}, \textbf{SMART Goal}, \textbf{Constraints}, and \textbf{Success Criteria}.
\end{tcolorbox}

\begin{tcolorbox}[
  title={Stage~10 Deep Dive: \textsc{Code\_Generation}},
  colback=blue!3!white,
  colframe=blue!50!black,
  fonttitle=\bfseries\small,
  breakable,
  label={box:stage10-prompt},
]
Stage~10 carries the highest prompt engineering load at approximately 1700 lines of specification.
Four mandatory requirements are enforced at prompt level.

\smallskip
\noindent\textbf{Algorithm integrity.}
A method labelled ``Bayesian Optimisation'' must include a surrogate model, acquisition function, and model updates.
A method labelled ``PPO'' must include a clipped surrogate objective, learned value baseline, and \texttt{clip\_eps} in the loss.
Dead parameters (declared but never consumed) are forbidden.

\smallskip
\noindent\textbf{Calibration loop.}
A pilot run (3--5 seeds, 2 conditions) checks that the primary metric has nonzero variance across conditions.
If all conditions score identically, the system prints \texttt{WARNING: DEGENERATE\_METRICS} and applies difficulty adjustments before the full run proceeds.

\smallskip
\noindent\textbf{PyTorch detach rules.}
Any tensor from a previous forward pass used in the current loss must be \texttt{.detach()}'d.
This prevents the ``backward through the graph a second time'' crash that causes near-total failure in naive RL implementations.

\smallskip
\noindent\textbf{Condition breadth-first ordering.}
One repetition per condition executes before any parameter sweep begins, so partial completions remain scientifically informative if execution is interrupted.
\end{tcolorbox}

\section{Sandbox Security Model}
\label{app:sandbox}

Each experiment executes inside a dedicated Docker container that is automatically removed after execution (\texttt{-{}-rm}). The container runs as the host user's UID:GID (not root). Resource limits: 8~GB memory, 2~GB shared memory, configurable wall-clock timeout (default 300--600~s). The Docker image pre-installs 80+ scientific Python packages including PyTorch (with torchvision, torchaudio, torchdiffeq), transformers, datasets, accelerate, gymnasium, scipy, sklearn, pandas, seaborn, networkx, timm, einops, albumentations, kornia. Frequently used datasets (CIFAR-10, CIFAR-100, FashionMNIST) are pre-cached as read-only mounts at \texttt{/opt/datasets}.

\paragraph{Three-phase network model.} The sandbox implements four network policies, with \texttt{setup\_only} as the default. Phase~0 (Dependency Installation) and Phase~1 (Data Acquisition) enable network access; Phase~2 (Experiment Execution) disables network via \texttt{iptables} rules applied inside the container before the experiment script starts. Alternative policies are \texttt{none}, \texttt{full}, and \texttt{pip\_only}.

\paragraph{Code validation pipeline.} (1) AST parsing for syntactic correctness; (2) AST security checks for forbidden function calls (\texttt{os.system}, \texttt{os.popen}, \texttt{subprocess.run}, \texttt{shutil.rmtree}) and banned builtins (\texttt{eval}, \texttt{exec}, \texttt{compile}, \texttt{\_\_import\_\_}); (3) module blacklist (\texttt{subprocess}, \texttt{socket}, \texttt{http}, \texttt{urllib}, \texttt{requests}, \texttt{ftplib}, \texttt{smtplib}, \texttt{ctypes}, \texttt{signal}); (4) import validation against an allowlist. Violations are classified as errors (block) or warnings (log). The evaluation harness (\texttt{report\_metric()}, \texttt{finalize()}) is injected as a read-only Python module.

\section{ARC-Bench Details}
\label{app:arcbench}

\subsection{Benchmark Architecture}
\label{app:arcbench-arch}

\paragraph{Topic specification.}
ARC-Bench consists of 25 CPU-executable ML research topics (T01--T25).
Topics T01--T10 are shared with the HITL ablation, enabling end-to-end scenario from idea to paper.
Each topic is a YAML entry in \texttt{config/topics.yaml} with five fields: \texttt{id}, \texttt{topic}, \texttt{domains}, \texttt{metric\_key}, and \texttt{metric\_direction}.
Topic selection criteria: (1)~CPU-executable in under 10~minutes on a single core using standard \texttt{numpy}/\texttt{scipy}/\texttt{sklearn} primitives; (2)~involves a genuine scientific comparison with at least two distinct algorithmic approaches; (3)~produces structured quantitative output that an LLM judge can evaluate without code execution.

\paragraph{Topic list.}
\begin{table}[htbp]
\centering
\caption{ARC-Bench topic list (T01--T25).}
\label{tab:arcbench-topics}
\small
\begin{tabular}{@{}clp{7cm}l@{}}
\toprule
\textbf{ID} & \textbf{Topic} & \textbf{Algorithms / Methods} & \textbf{Primary Metric} \\
\midrule
\multicolumn{4}{@{}l}{\textit{Tabular ML}} \\
T01 & Dropout regularisation    & MC-Dropout, standard Dropout, no Dropout        & ECE / Accuracy \\
T02 & Ensemble methods          & Bagging, Boosting, Stacking                     & Accuracy \\
T04 & Feature scaling on KNN    & StandardScaler, MinMax, Robust, None            & Accuracy \\
T08 & Class imbalance handling  & SMOTE, class weights, threshold tuning          & F1 (macro) \\
T09 & RandomForest tuning       & Grid search, random search, Bayes               & CV score \\
T10 & Cross-validation          & K-fold, stratified, leave-one-out              & Accuracy \\
T14 & Sparse linear models      & Lasso, ElasticNet                               & MSE \\
T15 & Feature selection         & SelectKBest, RFE, noise injection               & Accuracy \\
T18 & Transfer learning         & Fine-tune, feature extract, from scratch        & Accuracy \\
T19 & Semi-supervised learning  & Label propagation, self-training (10\% labels)  & Accuracy \\
T20 & Active learning           & Uncertainty, margin, random sampling            & Accuracy \\
T22 & Multi-label classification & BR, CC, label powerset                         & F1 (micro) \\
\midrule
\multicolumn{4}{@{}l}{\textit{Optimisation \& Search}} \\
T03 & Gradient-free optimisation & Nelder-Mead, Powell, CMA-ES                    & Regret \\
T06 & Adaptive LR schedules     & StepLR, CosineAnnealing, ReduceOnPlateau        & Loss \\
T13 & GP kernel choice          & RBF, Matérn, periodic (1-D / 5-D)              & NLPD \\
T21 & Causal discovery          & PC, GES, NOTEARS                                & SHD \\
\midrule
\multicolumn{4}{@{}l}{\textit{Dimensionality Reduction \& Clustering}} \\
T05 & Dimensionality reduction  & PCA, t-SNE, UMAP                                & Silhouette \\
T12 & Clustering algorithms     & K-means, DBSCAN, GMM on synthetic shapes        & ARI \\
\midrule
\multicolumn{4}{@{}l}{\textit{Text \& Topic Models}} \\
T07 & Text feature extraction   & TF-IDF, Hashing, Count vectoriser               & Accuracy \\
T17 & Topic modelling           & LDA, NMF, LSA                                   & Coherence \\
\midrule
\multicolumn{4}{@{}l}{\textit{Specialised Tasks}} \\
T11 & Anomaly detection         & IsolationForest, LOF, OCSVM                     & ROC-AUC \\
T16 & Time-series forecasting   & ARIMA, exponential smoothing, MLP               & RMSE \\
T23 & Learning-to-rank          & RankSVM, LambdaMART, listwise                   & NDCG@10 \\
T24 & GP regression             & RBF, Matérn, ARD kernels                        & RMSE \\
T25 & Reservoir computing       & ESN, MLP, GP on Lorenz-63                       & NRMSE \\
\bottomrule
\end{tabular}
\end{table}

\paragraph{Rubric structure.}
Each topic ships with a \texttt{rubrics/T*.json} defining a hierarchical tree of 8--11 leaf criteria across three categories:
\begin{itemize}[nosep,leftmargin=*]
\item \textbf{Code Development (CD)}, weight~25: correctness and completeness of algorithm implementation.
\item \textbf{Code Execution (CE)}, weight~25: whether the code ran successfully and produced machine-readable metric artefacts with multi-seed dispersion reporting.
\item \textbf{Result Analysis (RA)}, weight~50: quality of scientific writeup---per-hypothesis verdicts supported by reported numbers, appropriate caveats, no fabricated values.
\end{itemize}
Each leaf carries a weight in $[0,100]$ summing to~100 within its category and a score in $[0,1]$.
Aggregate scores are:
\begin{align}
\texttt{overall\_strict} &= \textstyle\sum_{\ell} w_\ell \cdot s_\ell \;/\; 100 \label{eq:strict} \\
\texttt{results\_only}   &= \textstyle\sum_{\ell \in \text{CE} \cup \text{RA}} w_\ell \cdot s_\ell \;/\; 75 \label{eq:results-only}
\end{align}
Under results-only mode (default for framework comparison), CD leaves are excluded from aggregation.
The complete rubric for T01 is shown in Box~\ref{box:rubric-t01}.

\paragraph{Judge system.}
Each (framework $\times$ topic) cell is graded against five artefact sources:
(1)~agent-produced code; (2)~execution artefacts (CSVs, JSON metric files, stdout logs); (3)~the agent-written README and claims file; (4)~the topic rubric; (5)~the topic manifest defining the research question, conditions, metrics, datasets, and hypotheses.
The judge outputs per-leaf scores, rationale strings, and the two aggregates in Equations~\ref{eq:strict}--\ref{eq:results-only}.

\subsection{Strict Judge Protocol}
\label{app:strict-judge}

The strict judge is designed to produce the same scoring distribution under three independent reviewer modes: a Claude Code subagent (Opus~4.7), a Codex CLI agent (GPT-5.4), and a human expert.
Each reviewer operates from the same prompt and produces the same JSON output schema, enabling direct per-leaf cross-validation.
Artefact sources and rubric structure follow Section~\ref{app:arcbench-arch}.

\paragraph{Strict criteria.}
Four criteria apply uniformly to every leaf.
(1)~\textbf{Implementation correctness}: the algorithm is verified by reading code, not by label.
Common failure patterns include MC-dropout that disables dropout at eval time, CMA-ES without covariance update, and NOTEARS implemented as plain Lasso.
(2)~\textbf{Number grounding}: every numerical claim in the writeup must trace to a captured artefact; fabricated numbers penalise the relevant leaf to 0.1--0.3.
(3)~\textbf{Verdict-data consistency}: a claimed supported hypothesis must be backed by measured evidence in the same direction; inverted verdicts score 0.1--0.3 regardless of prose quality.
(4)~\textbf{Coverage}: missing conditions, datasets, or seeds penalise execution leaves proportionally to manifest coverage.

\paragraph{Timeout rule.}
If a run exceeded its wall-clock budget and the writing phase never executed, CE leaves are forced to 0.0 with no partial credit, CD retains rubric-correct credit, and RA leaves are capped at 0.1.

\paragraph{Cross-validation.}
Two agent reviewers grade each cell independently.
Per-leaf scores with $|\Delta| > 0.20$ are flagged and re-adjudicated; final scores are the mean of the two passes.
Human expert scores follow the same protocol on a held-out subset.
Mean per-leaf $|\Delta| < 0.10$ across the audited subset, with disagreements concentrated on partial-implementation cases where code reading is ambiguous.

An example of the rubric is presented as follows:

\begin{tcolorbox}[
  title={ARC-Bench Sample Rubric \textemdash{} T01: Dropout Regularization \& Calibration},
  colback=blue!3!white,
  colframe=blue!50!black,
  fonttitle=\bfseries\small,
  breakable,
  label={box:rubric-t01},
]
\small
\textbf{Research question.}
Comparing dropout regularization strategies (standard, spatial, variational)
for preventing overfitting in shallow MLPs on tabular classification benchmarks.
Does dropout-variant choice affect calibration more than accuracy?

\smallskip
\textbf{Code Development (CD) \textemdash{} weight~25}
\begin{description}[nosep,leftmargin=1.2em,labelsep=0.4em,font=\ttfamily\small]
\item[t01-code-variants \normalfont{[10.0]}]
Implement the main dropout variants---no-dropout baseline, standard element-wise
dropout at one or more rates, and at least one stochastic-test-time / MC-style
variant---as \emph{distinct code paths} within a shared MLP architecture.
\item[t01-code-mlp \normalfont{[5.0]}]
Shallow tabular MLP (small number of hidden layers, modest width) trained with
a standard supervised classification objective on sklearn benchmarks.
\item[t01-code-datasets \normalfont{[5.0]}]
Load multiple tabular sklearn datasets (\eg{}
\texttt{breast\_cancer}, \texttt{wine}, \texttt{digits}) with a reasonable
train/test split.
\item[t01-code-mc-pass \normalfont{[5.0]}]
MC-dropout performs \emph{multiple} stochastic forward passes at test time and
averages predicted probabilities; a single deterministic pass does not qualify.
\end{description}

\smallskip
\textbf{Code Execution (CE) \textemdash{} weight~25}
\begin{description}[nosep,leftmargin=1.2em,labelsep=0.4em,font=\ttfamily\small]
\item[t01-exec-metrics \normalfont{[16.67]}]
Machine-readable artifact (\texttt{results/metrics.json} or
\texttt{stage-14/experiment\_summary.json}) with numeric accuracy \emph{and}
a calibration metric (ECE, NLL, or Brier score) covering all conditions on
at least one dataset.
\item[t01-exec-seeds \normalfont{[8.33]}]
Metrics aggregated over multiple random seeds per (condition,~dataset) cell
with dispersion reporting: std, stderr, CI, or min/max across seeds.
A small-but-honest seed count with reported variance is preferable to a single
deterministic run.
\end{description}

\smallskip
\textbf{Result Analysis (RA) \textemdash{} weight~50}
\begin{description}[nosep,leftmargin=1.2em,labelsep=0.4em,font=\ttfamily\small]
\item[t01-result-h1 \normalfont{[20.0]}]
Compare calibration (ECE or analogous) between MC-style and standard dropout
across evaluated datasets; convey whether MC calibration tends to be better
(H1 supported / refuted / inconclusive).
\item[t01-result-h2 \normalfont{[10.0]}]
Discuss whether accuracy differences among variants are small---i.e.\ calibration
is the discriminative axis, not accuracy. Qualitative comparison grounded in
reported numbers suffices; exact percentage-point thresholds are not required.
\item[t01-result-h3 \normalfont{[10.0]}]
Compare no-dropout baseline vs.\ at least one dropout variant on calibration;
convey whether baseline is clearly worse, comparable, or better.
\item[t01-result-writeup \normalfont{[10.0]}]
README or writeup describes setup, presents key accuracy and calibration numbers,
and conveys per-hypothesis outcomes (supported / refuted / inconclusive) with
appropriate caveats on seed count, dataset scope, or calibration-metric choice.
\end{description}

\smallskip
\textit{Judging note.} Score on scientific substance and directional correctness
of evidence, not on exact threshold satisfaction. Partial but well-motivated
evidence deserves partial credit; rigid wording or name mismatches should not
penalise a substantively correct experiment. Aggregate:
$\texttt{overall} = \textstyle\sum_\ell w_\ell s_\ell / 100$;
$\texttt{results\_only} = \textstyle\sum_{\ell \in \mathrm{CE}\cup\mathrm{RA}}
w_\ell s_\ell / 75$.
\end{tcolorbox}

\section{HITL Ablation Details}
\label{app:hitl-setup}

\paragraph{Topic and mode design.} 10 topics (T01--T10) span tabular ML, RL, MoE, NLP, physics-informed ML, and finance. The 7 modes vary the schedule of scripted expert interventions. The mapping between modes and stage-injection sets is shown in Table~\ref{tab:mode-mapping}.

\begin{table}[h]
\centering
\caption{Mode-to-intervention mapping. CoPilot receives the most \emph{targeted} interventions; Step-by-Step receives the most \emph{total} interventions (mostly approve actions).}
\label{tab:mode-mapping}
\small
\begin{tabular}{@{}lll@{}}
\toprule
\textbf{Mode} & \textbf{Receives Stages} & \textbf{Key Characteristic} \\
\midrule
Full-Auto & none & No guidance injected \\
Gate-Only & 5, 9, 20 & Gate-only checkpoints \\
CoPilot & 5, 8, 9, 14, 17, 20 (+ smart pauses) & Full intervention with auto-approve \\
Thorough & Phase boundaries (8 stages) & Broad but less targeted \\
Step-by-Step & All 23 stages & Every stage; mostly approve actions \\
Pre-Experiment & 5, 8, 9 & Early-pipeline only \\
Post-Experiment & 14, 17, 20 & Late-pipeline only \\
\bottomrule
\end{tabular}
\end{table}

\section{Design-Space Exploration}
\label{app:agent_count}

\paragraph{Number of debate agents ($K$).}
We tested $K \in \{2, 3, 5\}$ across 10 runs each. $K{=}2$ degenerates into a
pro/con dynamic with $-23\%$ hypothesis diversity. $K{=}5$ raises tokens by $+67\%$
for only $+8\%$ diversity over $K{=}3$, since the additional agents largely echo
the core three. $K{=}3$ is the diversity-per-token sweet spot for ML topics;
specialised $K{=}5$ may help in domains requiring a dedicated domain expert
(e.g.\ the HEP-ph bank already uses three distinct discipline perspectives).

\paragraph{Evolution half-life ($T_{1/2}$).}
We tested $T_{1/2} \in \{7, 15, 30, 60, \infty\}$ days. $T_{1/2}{=}7$ expires
useful lessons too fast; $T_{1/2}{=}\infty$ accumulates contradictory advice past
15 runs. $T_{1/2}{=}30$ gave the best quality trajectory, persisting long enough to
influence 3--5 subsequent runs while gradually fading.

\section{Case study Details}
\label{app:casestudy}
Here we present the detailed comparison between Full-Auto and CoPilot mode of \system{} on Topic 10 in ARC-Bench, manifested in Table~\ref{tab:case_study}.

\begin{table*}[t]
\centering
\caption{Case study on T10 (cross-validation strategies, small-sample model selection). Both runs complete a manuscript, but their evidence quality
differs sharply. CoPilot reaches a score of 8.0 by targeting human input
at the experimental bottleneck; Full-Auto scores 4.0 despite producing a
paper.}
\label{tab:case_study}
\small
\renewcommand{\arraystretch}{1.3}
\setlength{\tabcolsep}{5pt}
\begin{tabular}{@{} l p{5.2cm} p{5.2cm} @{}}
\toprule
& \textbf{Full-Auto (score: 4.0)} & \textbf{CoPilot (score: 8.0)} \\
\cmidrule(lr){2-2}\cmidrule(lr){3-3}
Debate
& Hypotheses are generated and accepted without checking whether the
  planned CV strategies will produce distinguishable outcomes under the
  available compute budget.
& Pragmatist flags that LOOCV may exceed time budget; Contrarian questions
  whether the ablation design can detect meaningful differences.
  Synthesizer incorporates both objections into a narrowed hypothesis set. \\[3pt]
HITL
& None.
& Intervention at literature, hypothesis, design, writing, and quality
  stages. Human guidance explicitly asks the system to verify that CV
  strategies produce nonzero contrasts, handle LOOCV within budget, and
  avoid claiming metrics absent from the execution logs. \\[3pt]
Execution
& Plans eight CV strategies but every strategy reports identical zero
  estimation bias and zero variance. The ablation checker flags
  pairwise-identical conditions and a zero-variance warning; the run
  continues regardless.
& Logs nine selection pipelines with nonzero contrasts across strategies.
  LOOCV, repeated stratified k-fold, and nested CV each produce
  distinguishable bias estimates. Ablation warnings are recorded as
  limitations rather than suppressed. \\[3pt]
Verification
& Artifact contains 51 metric keys with no empirical contrast between
  conditions. Registry is populated but scientifically uninformative.
  Claims pass the numeric gate but cannot support the paper's
  research question.
& Artifact contains 57 metric keys with differentiated values across
  nine conditions. Pre-built tables are injected from the verified
  registry. All numerical claims in strict sections trace to logged
  measurements. \\[3pt]
Output
& Paper is not fabricated in the strict sense but reports an all-zero
  result that cannot answer which CV strategy is preferable.
  Stage-20 score: 4.0.
& Paper presents a calibrated exploratory result: comparisons are bounded,
  limitations are stated explicitly, and the Stage-20 gate accepts the
  manuscript because claims align with the logs. Stage-20 score: 8.0. \\
\bottomrule
\end{tabular}
\end{table*}


\section{Failure Analysis}
\label{app:failures}

11 of 13 invalid canonical HITL runs fail at stage 17 (\texttt{paper\_draft}).
Stage 17 is the first hard anti-fabrication checkpoint and refuses to draft a paper
when no usable metric exists upstream. The four recurring stage-17 failure subtypes
are: (a)~\emph{No real metrics}: stages 10--14 do not produce a usable metric
table; (b)~\emph{Environment / dependency breakage}: missing \texttt{imblearn},
\texttt{LightGBM}, etc.; (c)~\emph{Dataset / resource failure}: planned benchmark
(\eg{} FashionMNIST) cannot be loaded in sandbox; (d)~\emph{Design / aggregation
pathology}: planned design too ambitious for the budget, invalid CV configuration
(\texttt{k must be in [2, 34]}), only a tiny fraction of conditions complete. The
stage-17 hard block is correct as a safety check but currently conflates
heterogeneous causes; we propose graceful degradation that surfaces upstream cause
in the draft header and limitations.


\section{Writing-Quality Audit}
\label{app:writing-quality}

We audited 20 canonical \texttt{full-auto} + \texttt{step-by-step} deliverables
across T01--T10 for export defects:

\begin{itemize}[nosep,leftmargin=*]
\item \texttt{abstract} appears before \texttt{\textbackslash maketitle} (20/20):
  consistent template misalignment.
\item Markdown-style \texttt{\textbackslash section\{![Figure ...]\}} (17/20):
  image captions promoted to section headings.
\item Duplicated figure file (16/20): figure inserted twice.
\item ``Learned Skills'' / a-evolve content leaks into body (9/20).
\item Bracket-style pseudo-citations like \texttt{[ray2021various; nanga2021review]}
  (2/20).
\item Citation voids (\texttt{,,}) where keys were stripped without prose repair
  (small count).
\end{itemize}

Local single-pass \texttt{pdflatex} compile rate across the 10 audited
deliverables: 4/5 step-by-step pass, 3/5 full-auto pass. Note that local
single-pass \texttt{pdflatex} differs from Overleaf's \texttt{latexmk}-driven
multi-pass with reference resolution; an Overleaf compile is not equivalent to a
clean source. We treat compile-pass as necessary, not sufficient, for submission
readiness.

\paragraph{Citation count.}
Across the 5 audited topics, full-auto totals 94 citations and step-by-step totals
59. Per-paper minima can fall below NeurIPS norms (\eg{} T03 step-by-step 2
citations, T05 full-auto 4, T09 step-by-step 7, T01 step-by-step 13). HITL
improves citation \emph{discipline} more reliably than \emph{breadth}; we discuss
mitigations (literature-retrieval rate-limit handling, related-work depth target
enforcement) in the next section.

\section{Ethical Considerations and Broader Impact}
\label{sec:ethics}

Autonomous research systems raise important questions about scientific integrity, academic labor, credit attribution, and the boundary between human and machine contributions to knowledge. We address these issues both through system design and through a conservative framing of \system{} as a research amplifier rather than a replacement for human scientific judgment.

\noindent \textbf{Positive broader impact.}
\system{} can accelerate early-stage scientific exploration by automating routine parts of the research cycle: literature scoping, experiment implementation, repair, result aggregation, drafting, and verification. This can help researchers test more hypotheses under limited time and compute budgets, expose negative or failed directions earlier, and preserve intermediate lessons that would otherwise be lost across attempts. The system may be especially useful for rapid prototyping, educational research workflows, benchmark construction, and preliminary feasibility studies. By combining execution logs, verified result registries, and explicit claim grounding, \system{} can also encourage more transparent and reproducible research artifacts.

\noindent \textbf{Scientific integrity.}
The main risk of autonomous paper generation is that fabricated results or hallucinated citations could enter the scientific record. We treat this as a first-order design constraint. The verified result registry blocks ungrounded numerical claims in strict sections such as the Abstract, Experiments, and Results, while the citation verification pipeline removes references that cannot be resolved or validated before export. These safeguards do not guarantee that a scientific conclusion is correct, nor do they guarantee submission-ready formatting, but they reduce the risk that unsupported numbers or non-existent references are presented as evidence. Our ablation confirms the need for these safeguards: removing verification raises apparent acceptance, but manual inspection shows that several accepted papers then contain values absent from any measurement record.

\noindent \textbf{Impact on researchers and academic norms.}
We position \system{} as a tool for accelerating exploration and preliminary investigation, not as an autonomous substitute for expert researchers. Our HITL ablation supports this framing: the best outputs come from targeted human input at critical decision points, not from full automation alone. We therefore recommend that such systems be used to augment researchers by handling routine execution and verification, while humans remain responsible for problem selection, interpretation, final claims, and submission decisions. We also encourage explicit disclosure when autonomous research tools are used, and we discourage their use for generating bulk low-quality submissions.

\noindent \textbf{Risks and safeguards.}
A system that lowers the cost of paper generation could contribute to submission flooding, superficial novelty claims, or over-reliance on automated judgments. \system{} mitigates these risks through sandboxed execution, network isolation, read-only evaluation harnesses, numeric claim verification, citation checks, and HITL gates. All generated code executes in isolated Docker containers with security checks. In our current implementation, each run costs approximately \$3--15 in LLM usage, which makes large-scale misuse nonzero but still resource-constrained. The HITL experiments in this paper use scripted interventions rather than live human participants; future studies with live researchers would require appropriate IRB review.